\pdfoutput=1
\documentclass[11pt,a4paper]{article}

\usepackage[utf8]{inputenc}
\usepackage[T1]{fontenc}
\usepackage{amsmath,amssymb,amsthm}
\usepackage{graphicx}
\usepackage{hyperref}
\usepackage[numbers]{natbib}
\usepackage{booktabs}
\usepackage{tikz}
\usetikzlibrary{positioning, arrows.meta, shapes.geometric, calc, matrix}
\usepackage{listings}
\usepackage{xcolor}
\usepackage{lmodern}
\definecolor{headcolor}{HTML}{000000}
\definecolor{dimcolor}{gray}{0.6}
\newcommand{\head}[1]{{\texttt{\bfseries\color{headcolor}#1}}}
\newcommand{\tdim}[1]{{\texttt{\color{dimcolor}#1}}}
\usepackage{fancyvrb}
\usepackage{float}
\usepackage{subcaption}
\usepackage[margin=1in]{geometry}

\newtheorem{theorem}{Theorem}

\newtheorem{proposition}[theorem]{Proposition}
\newtheorem{definition}[theorem]{Definition}
\newtheorem{remark}{Remark}
\newtheorem{example}{Example}

\newcommand{\R}{\mathbb{R}}

\newcommand{\relu}{\mathrm{ReLU}}
\newcommand{\clamp}{\operatorname{clamp}}
\newcommand{\sat}{\sigma}

\title{A PyTorch Library of Turing-Complete Neural Networks}

\author{
Jonathan Bates \\
jonrbates.ai \\
\texttt{github.com/jonrbates/turing}
}

\date{}

\begin{document}
\maketitle

\begin{abstract}
We present a PyTorch package that compiles neural networks
and their weights from Turing machine descriptions,
producing models that exactly simulate the specified machine
without any training.
Given a transition function and a set of terminal states,
the package constructs a model whose forward pass
corresponds to one step of the Turing machine.
Two architectures are implemented, each realizing a different
theoretical result: (1)~a transformer with self-attention,
cross-attention, and feedforward layers based on
Wei, Chen, and Ma~(2021), and (2)~a recurrent network based on
Siegelmann and Sontag~(1995) that encodes the stack in a
Cantor set.
We develop the constructions from first principles, showing how
ReLU networks implement Boolean circuits (AND, OR, NOT, XOR gates
and their composition into DNF formulas and binary
adders) and how hard attention implements positional lookup on
the tape.
The package serves as a concrete, runnable reference for the
symbolic--neural bridge, and as a foundation for future work on
the stability of constructed solutions under gradient-based
optimization.
Code is available at \url{https://github.com/jonrbates/turing}.
\end{abstract}

\section{Introduction}
\label{sec:intro}

Can a neural network simulate a Turing machine?
The question has been answered affirmatively multiple times in the
theoretical literature: by Siegelmann and Sontag~\cite{siegelmann1992,siegelmann1995}
using recurrent networks, by P\'{e}rez et al.~\cite{perez2021}
using transformers with hard attention, and by Wei, Chen, and
Ma~\cite{wei2021} using a finite-precision transformer
with binary positional encoding.
More recently, Li and Wang~\cite{li2025} showed that even
constant bit-size transformers are Turing complete, given a
sufficiently long context window and chain-of-thought steps.
These results give explicit constructions, but they live on paper:
the gap between a proof specifying weight matrices and a
runnable implementation remains.

In this paper, we describe a PyTorch package that closes that gap.
Given a Turing machine specified by its transition function and
terminal states, the package constructs a neural network and
\emph{specifies} all weights and biases so that the
network's forward pass exactly implements one step of the machine.
No training is required: the network is correct by construction.
(For the WCM21 transformer, weights are set by direct construction
from the transition table; for the SS95 recurrent network,
a least-squares solution determines the weights from the
configuration space.)

The compilation from machine description to neural network
proceeds through a hierarchy of components:
\begin{enumerate}
\item \textbf{Logic gates.} ReLU networks compute AND,
  OR, NOT, and XOR on binary-valued subsets of the
  state vector.
\item \textbf{Transition function.} The finite lookup table
  $\delta$ is realized as a two-layer DNF circuit: one hidden
  neuron per state-symbol pair.
\item \textbf{Binary arithmetic.} Half-adders and full adders,
  composed from the above gates, form a ripple-carry adder
  that updates the head position.
\item \textbf{Positional lookup.} Hard-attention layers with
  binary key-query matching retrieve symbols from the tape
  history and the original input.
\item \textbf{Symbol assembly.} Feedforward layers
  select the correct symbol at the new head position
  by priority: last write, original tape, or empty.
\end{enumerate}

Every weight in the resulting network has a known, interpretable
purpose.
This interpretability is a property of the \emph{construction process},
not of post-hoc analysis, a point we return to in the discussion.

Two simulation architectures are provided:
\begin{itemize}
\item \textbf{WCM21} (Wei, Chen \& Ma, 2021): an encoder-decoder transformer
  in which the encoder stores the initial tape and the decoder
  autoregressively tracks the computation history.
  Self-attention retrieves past writes; cross-attention reads
  the original tape; feedforward layers implement the transition
  function and binary arithmetic.
\item \textbf{SS95} (Siegelmann \& Sontag, 1992; 1995): a recurrent network that
  reformulates the Turing machine as a stack machine and encodes
  each stack as a single rational number using a Cantor-set encoding.
  Two versions are implemented: a 4-layer network with explicit
  pipeline stages, and a 1-layer ``real-time'' variant.
\end{itemize}

\paragraph{Contributions.}
\begin{enumerate}
\item A self-contained exposition of how ReLU networks realize
  Boolean circuits, building from individual gates through DNF
  to binary adders.
\item A constructive implementation of two Turing machine simulation
  results as a PyTorch package with weights specified from the
  machine description.
\item An open-source codebase suitable for studying the stability
  of constructed solutions under perturbation and
  gradient-based fine-tuning.
\end{enumerate}

\paragraph{Related work.}
P\'{e}rez et al.~\cite{perez2021} gave an explicit construction
of a transformer with hard attention that simulates any Turing
machine, using a single-layer encoder and three-layer decoder.
Their construction operates over rationals of arbitrary precision;
when multiple positions tie for the maximum attention score,
the hardmax produces fractions like $1/t$, requiring precision
to grow with the sequence length.

Wei, Chen, and Ma~\cite{wei2021} avoid this by using
\emph{binary positional encoding}. Head positions and step
numbers are represented as fixed-width binary vectors, and
position lookup is performed by matching binary bits via
attention scores that never produce fractional ties.
This is the key modification that makes the construction
implementable at finite precision. The number of parameters
is polynomial in the alphabet size $|\Gamma|$, state space size $|Q|$,
and $\log T$, rather than growing with the input length.
The WCM21 implementation follows the Wei-Chen-Ma construction
directly, translating Claims C.4--C.6 and Lemmas C.1, C.2,
C.7, C.8 into PyTorch modules; the binary encoding is realized
by the ripple-carry adder and binary search layers.

The next line of work introduced \emph{chain-of-thought (CoT)}
as a computational resource for transformers.
Li et al.~\cite{li2024cot} proved that constant-depth,
constant-precision transformers augmented with $O(\log t)$
chain-of-thought steps can solve problems requiring serial
computation, and Merrill and Sabharwal~\cite{merrill2024cot}
characterized the expressive power of such CoT-augmented
transformers, showing that polynomial CoT steps yield the
power of polynomial-time Turing machines.
These results formalize the intuition that intermediate
generation steps can serve as a work tape.

Li and Wang~\cite{li2025} pushed this further, proving that even
constant bit-size transformers (fixed precision and fixed
parameter count) are Turing complete when given sufficient
context window length, using Post machine simulation with
chain-of-thought steps.

\begin{table}[H]
\centering
\small
\caption{Resource requirements for neural Turing-completeness
  constructions.
  $t$ and $s$ denote the worst-case runtime and space
  of the simulated machine over inputs of length $n$.
  Transformer rows adapted from Li and Wang~\cite{li2025}, Table~1.}
\label{tab:tc_comparison}
\begin{tabular}{@{}lcccc@{}}
\toprule
 & Precision & Dim.\ & Window & CoT/step \\
\midrule
\textbf{Siegelmann \& Sontag~\cite{siegelmann1992,siegelmann1995} (SS95)}
  & $O(s)$ & $O(|Q|)$ & --- & --- \\
P\'{e}rez et al.~\cite{perez2021}
  & $O(\log t)$ & $O(1)$ & $n{+}t$ & 1 \\
Bhattamishra et al.~\cite{bhattamishra2020}
  & Unbounded & $O(1)$ & $n{+}t$ & 1 \\
\textbf{Wei et al.~\cite{wei2021} (WCM21)}
  & $O(1)$ & $O(|Q||\Gamma|{+}\log t)$ & $n{+}t$ & 1 \\
Li et al.~\cite{li2024cot}
  & $O(1)$ & $O(\log t)$ & $O(t \log t)$ & $O(\log t)$ \\
Merrill \& Sabharwal~\cite{merrill2024cot}
  & $O(\log t)$ & $O(1)$ & $n{+}t$ & 1 \\
Li \& Wang~\cite{li2025}
  & $O(1)$ & $O(1)$ & $s$ & $s$ \\
\bottomrule
\end{tabular}
\end{table}

Smolensky et al.~\cite{smolensky2024} developed PSL
(Production System Language), a high-level language for symbolic
programs that compiles into transformer weights using Tensor
Product Representations.
PSL operates at the level of production rules over structured
working memory; the constructions here expose the lower-level
circuit primitives (gates, adders, hard attention) that realize
the computation.

The Neural Turing Machine~\cite{graves2014} and related
differentiable memory architectures are inspired by and
inductively biased to emulate computer architecture.
The approach here instead compiles any given
program into fixed weights.

\section{Preliminaries}
\label{sec:prelim}

\subsection{Turing Machines}

\begin{definition}[Turing Machine]
A Turing machine is a tuple $M = (Q, \Gamma, \delta, q_0, F)$ where
$Q$ is a finite set of states,
$\Gamma$ is a finite tape alphabet,
$\delta: Q \times \Gamma \to Q \times \Gamma \times \{-1, +1\}$
is the (partial) transition function,
$q_0 \in Q$ is the initial state, and
$F \subseteq Q$ is the set of terminal (halting) states.
Given a state $q$ and a symbol $a$ under the head,
$\delta(q, a) = (q', a', d)$ specifies the next state $q'$,
the symbol $a'$ to write, and the direction $d$ to move the head.
\end{definition}

A \emph{configuration} at time $t$ is the triple
$(q_t, h_t, \tau_t)$: the current state $q_t \in Q$,
the head position $h_t \in \{0, \ldots, n-1\}$, and
the tape contents $\tau_t \in \Gamma^n$.
The machine halts when $q_t \in F$.

\begin{example}[Balanced Parentheses]
\label{ex:bp}
Consider the language of balanced parenthesis strings.
We define a Turing machine with states $Q = \{I, R, M, V, T, F\}$,
alphabet $\Gamma = \{B, E, (, ), *\}$, initial state $q_0 = I$,
and terminal states $\{T, F\}$ (``True'' and ``False'').
The tape is initialized with sentinels $B$ and $E$ flanking
the input, e.g., \texttt{B()((()(()))())E}.
The transition function is given in Table~\ref{tab:bp_delta}.

The machine scans right in state $R$ until it finds a close
parenthesis; switches to state $M$ and scans left to find
a matching open parenthesis (replacing both with $*$);
then resumes scanning right.
State $V$ verifies that no unmatched symbols remain.
\end{example}

\begin{table}[htbp]
\centering
\caption{Transition function $\delta$ for the balanced parentheses machine.}
\label{tab:bp_delta}
\begin{tabular}{@{}ccccc@{}}
\toprule
State & Read & Next State & Write & Move \\
\midrule
$I$ & $B$ & $R$ & $B$ & $+1$ \\
$R$ & $($ & $R$ & $($ & $+1$ \\
$R$ & $)$ & $M$ & $*$ & $-1$ \\
$R$ & $*$ & $R$ & $*$ & $+1$ \\
$R$ & $E$ & $V$ & $E$ & $-1$ \\
$M$ & $B$ & $F$ & $*$ & $-1$ \\
$M$ & $($ & $R$ & $*$ & $+1$ \\
$M$ & $*$ & $M$ & $*$ & $-1$ \\
$V$ & $($ & $F$ & $*$ & $-1$ \\
$V$ & $*$ & $V$ & $*$ & $-1$ \\
$V$ & $B$ & $T$ & $B$ & $+1$ \\
\bottomrule
\end{tabular}
\end{table}

\begin{figure}[H]
\centering
\begin{tabular}[t]{@{}l@{\quad}l@{}}
$I$ & \head{B}\tdim{()E} \\
$R$ & \tdim{B}\head{(}\tdim{)E} \\
$R$ & \tdim{B(}\head{)}\tdim{E} \\
$M$ & \tdim{B}\head{(}\tdim{*E} \\
$R$ & \tdim{B*}\head{*}\tdim{E} \\
$R$ & \tdim{B**}\head{E} \\
$V$ & \tdim{B*}\head{*}\tdim{E} \\
$V$ & \tdim{B}\head{*}\tdim{*E} \\
$V$ & \head{B}\tdim{**E} \\
$T$ & \tdim{B}\head{*}\tdim{*E} \\
\end{tabular}
\caption{Execution trace on input \texttt{B()E}.
The head scans right in state $R$,
matches parentheses in state $M$, and verifies in state $V$.
Matched pairs are replaced with \texttt{*}.
Head symbol \head{highlighted}.}
\label{fig:tape}
\end{figure}

\subsection{Notation and Activations}

We write $\relu(x) = \max(x, 0)$ for the rectified linear unit.
When inputs are restricted to $\{0, 1\}$, the constructions in
this paper are designed so that all intermediate values remain
in $\{0, 1\}$ under $\relu$.
The \emph{saturated linear} activation
$\sat(x) = \clamp(x,\, 0,\, 1)$, which explicitly clamps outputs
to $[0, 1]$, is used in the SS95 construction
(\S\ref{sec:ss}) where Cantor-set encoded values require
bounded activations.

\section{Boolean Circuits in ReLU Networks}
\label{sec:circuits}

A fundamental observation underlying the constructions is that
networks with piecewise-linear activations can exactly
implement Boolean logic.
We develop this systematically.

\subsection{Elementary Gates}

\begin{proposition}[NOT]
\label{prop:not}
For $x \in \{0, 1\}$:
$\neg x = \relu(1 - x)$.
\end{proposition}

\begin{proposition}[AND]
\label{prop:and}
For $x_1, x_2 \in \{0, 1\}$:
$x_1 \wedge x_2 = \relu(x_1 + x_2 - 1)$.
This generalizes: $\bigwedge_{i=1}^{k} x_i = \relu\!\left(\sum_{i=1}^{k} x_i - (k-1)\right)$.
\end{proposition}

\begin{proposition}[NOR and OR]
\label{prop:or}
$\mathrm{NOR}(x_1, x_2) = \relu(1 - x_1 - x_2)$ and
\begin{equation*}
\mathrm{OR}(x_1, x_2) = \mathrm{NOT}(\mathrm{NOR}(x_1, x_2)) = \relu(1 - \relu(1 - x_1 - x_2)).
\end{equation*}
\end{proposition}

\begin{remark}
AND and NOR are both single-neuron operations with the same
structure (a weighted sum followed by $\relu$), differing only
in the sign of the weights and the bias.
This allows them to be computed \emph{simultaneously} in one layer
by writing to different positions.
OR and NAND are then obtained by negation in the next layer.
\end{remark}

\begin{proposition}[XOR]
\label{prop:xor}
$x_1 \oplus x_2$ is not linearly separable and cannot be computed by a single neuron. The implementation uses:
\begin{equation}
x_1 \oplus x_2 = \mathrm{AND}(\mathrm{NAND}(x_1, x_2),\; \mathrm{OR}(x_1, x_2)).
\end{equation}
\end{proposition}

\subsection{Embedding Gates in Weight Matrices}

A key implementation technique is that gates operating on
a \emph{subset} of dimensions can be embedded in a large
weight matrix that acts as the identity on all other dimensions.
Starting from $W = I$, we modify the rows and columns
corresponding to the gate's inputs and outputs.
This allows multiple gates to be applied in parallel, and
circuits to be composed by sequential matrix-$\relu$ layers.

For example, to compute $\mathrm{AND}(x_i, x_j)$ and store
the result at position $i$ while passing all other dimensions
through unchanged:
\begin{equation}
W = I, \quad W_{ij} = 1, \quad b_i = -1,
\end{equation}
so that $[\relu(Wx + b)]_i = \relu(x_i + x_j - 1)$ and
$[\relu(Wx + b)]_k = x_k$ for $k \neq i$.

\subsection{DNF Realization and the Transition Function}
\label{sec:dnf}

Any Boolean function $f: \{0,1\}^n \to \{0,1\}$ can be written in
disjunctive normal form (DNF):
$f(x) = \bigvee_{j} C_j(x)$,
where each clause $C_j$ is a conjunction of literals.

\begin{theorem}[DNF Realization]
\label{thm:dnf}
Any DNF with $m$ clauses over $n$ variables can be computed
by a three-layer ReLU network: Layer~1 computes each clause
via AND ($m$ hidden neurons), Layer~2 computes NOR of the
clause outputs, and Layer~3 negates to produce OR.
When the clauses are mutually exclusive (at most one is active),
Layer~2 can compute OR directly by summation,
reducing the network to two layers.
\end{theorem}

The Turing machine transition function $\delta$ is a finite
lookup table mapping $(q, a)$ pairs to $(q', a', d)$ triples.
With one-hot encodings, each table entry is a clause:
``state is $q$'' AND ``symbol is $a$''.
The first layer appends $|Q| \cdot |\Gamma|$ detector dimensions
(one per state-symbol pair), each a 2-input AND with bias $-1$.
After $\relu$, exactly one detector fires.
The second layer routes the outputs: for each entry
$\delta(q, a) = (q', a', d)$, it sets a 1 in the output
column for next state $q'$, write symbol $a'$, and direction $d$.
Position-encoding bits (\texttt{pos1}, \texttt{pos2}) are
passed through unchanged via the identity block.

In the package, this is the \texttt{Transition} module:
$W_1 \in \R^{(w + sg) \times w}$ and $W_2 \in \R^{w \times (w + sg)}$.

\section{Binary Arithmetic}
\label{sec:adders}

To update the head position, we must add or subtract 1 from
a binary-encoded integer.
This requires binary addition circuits built from the gates
of \S\ref{sec:circuits}.

\subsection{Half Adder}

A half adder takes two bits $a, b \in \{0, 1\}$ and produces
a sum $s = a \oplus b$ and a carry $c = a \wedge b$.
The implementation uses three layers,
each embedded in a $d$-dimensional identity matrix,
operating on positions $i$ (overwritten with sum) and
$j$ (overwritten with carry):

\begin{enumerate}
\item Compute $\mathrm{AND}(x_i, x_j)$ at position $i$ and
  $\mathrm{NOR}(x_i, x_j)$ at position $j$ simultaneously.
\item Negate both: $\mathrm{NAND} = \mathrm{NOT}(\mathrm{AND}(x_i, x_j))$ at position $i$,
  $\mathrm{OR} = \mathrm{NOT}(\mathrm{NOR}(x_i, x_j))$ at position $j$.
\item Combine:
  $s = x_i \oplus x_j = \mathrm{AND}(\mathrm{NAND}, \mathrm{OR})$ at position $i$;
  $c = \mathrm{NOT}(\mathrm{NAND})$ at position $j$.
\end{enumerate}

AND and NOR share the first layer because they have the same
computational structure (weighted sum plus $\relu$) with
opposite signs.
Listing~\ref{lst:halfadder} shows the complete PyTorch
implementation.
Each layer starts from the $d \times d$ identity matrix
and modifies only the rows for positions $i$ and $j$.
All other dimensions pass through unchanged, so the half
adder can be embedded in an arbitrarily large state vector.

\begin{lstlisting}[caption={The half adder. Three linear-ReLU layers
compute $\mathrm{sum} = x_i \oplus x_j$ at position $i$
and $\mathrm{carry} = x_i \wedge x_j$ at position $j$.},
label={lst:halfadder},float=htbp]
# Layer 1: AND at position i, NOR at position j
W1 = torch.eye(d_in)
b1 = torch.zeros(d_in)
W1[i, j] = 1;  b1[i] = -1   # relu(x_i + x_j - 1) = AND
W1[j, i] = -1; W1[j, j] = -1; b1[j] = 1
                               # relu(1 - x_i - x_j) = NOR

# Layer 2: NAND at position i, OR at position j
W2 = torch.eye(d_in)
b2 = torch.zeros(d_in)
W2[i, i] = -1; b2[i] = 1    # relu(1 - AND) = NAND
W2[j, j] = -1; b2[j] = 1    # relu(1 - NOR) = OR

# Layer 3: XOR at position i, AND at position j
W3 = torch.eye(d_in)
b3 = torch.zeros(d_in)
W3[i, j] = 1;  b3[i] = -1   # relu(NAND + OR - 1) = XOR
W3[j, i] = -1; W3[j, j] = 0; b3[j] = 1
                               # relu(1 - NAND) = AND
\end{lstlisting}

\subsection{Full Adder}

A full adder takes three bits $a$, $b$, $c_{\mathrm{in}}$ and
produces $s = a \oplus b \oplus c_{\mathrm{in}}$ and
$c_{\mathrm{out}} = (a \wedge b) \vee (c_{\mathrm{in}} \wedge (a \oplus b))$.
It cascades two half-adders followed by two layers that compute
the OR of the intermediate carries (via NOR then NOT).

In the WCM21 simulator, a chain of $\lceil \log_2 T \rceil$
full adders forms a ripple-carry adder.
The \texttt{PreprocessForAdder} module sets up the operands:
for a right move ($d = +1$), the addend is the binary
representation of 1 (a single 1-bit in the least significant
position);
for a left move ($d = -1$), the addend is all ones,
implementing subtraction via two's complement.
The direction is read from \texttt{scr5}, which was set by
the transition layer.

\section{The WCM21 Construction}
\label{sec:wcm}

We now describe the full simulation architecture based on
Wei, Chen, and Ma~\cite{wei2021}.
The network is an encoder-decoder transformer whose
weights are specified from the machine description.

\subsection{Architecture Overview}

The \textbf{encoder} processes the initial tape once.
Each of the $n$ tape positions is encoded as a row of a
matrix $E \in \R^{n \times w}$, containing a one-hot symbol
encoding and a binary position index.
$E$ is fixed throughout the simulation.

The \textbf{decoder} maintains a growing sequence of state
vectors $H = [h_0, h_1, \ldots, h_{t-1}]$, one per
simulation step.
Each vector $h_t \in \R^w$ encodes the machine state, the
symbol at the current position, the step number, the head
position, and scratch space for intermediate computation.
After each step, a position embedding $\beta(t)$ encoding
the step number is added to the new vector before it is
appended to $H$.

The embedding dimension $w$ is partitioned into named slices:

\begin{center}
\begin{tabular}{ll}
\toprule
Slice & Contents \\
\midrule
\texttt{st} & One-hot machine state ($|Q|$ dims) \\
\texttt{sym1} & One-hot current symbol ($|\Gamma|$ dims) \\
\texttt{sym2} & One-hot written symbol ($|\Gamma|$ dims) \\
\texttt{pos1} & Binary step number ($\lceil \log_2 T \rceil$ dims) \\
\texttt{pos2} & Binary head position ($\lceil \log_2 T \rceil$ dims) \\
\texttt{pos3} & Binary new head position ($\lceil \log_2 T \rceil$ dims) \\
\texttt{scr1--scr5} & Scratch: last-written symbol, initial symbol, \\
& \quad binary search state, visit flags, direction \\
\bottomrule
\end{tabular}
\end{center}

For the balanced parentheses example (6 states, 5 symbols,
$T = 100$), we have $\lceil \log_2 100 \rceil = 7$, giving
$w = 6 + 2(5) + 3(7) + 22 = 59$.

\subsection{One Simulation Step}

Each step applies the following pipeline to the most recent
decoder vector, producing a new vector that is appended to $H$.
We reference the corresponding claims and lemmas
from~\cite{wei2021}.

\begin{enumerate}
\item \textbf{Transition} (feedforward, Lemma C.1).
  Two linear layers with $\relu$ implement the
  transition function $\delta$ as a DNF circuit
  (\S\ref{sec:dnf}).
  Reads \texttt{st} and \texttt{sym1}; writes next state to
  \texttt{st}, written symbol to \texttt{sym2}, and direction
  to \texttt{scr5}.

\item \textbf{Preprocess for Adder} (linear, Lemma C.2).
  Copies \texttt{pos2} into \texttt{pos3} and extends
  the vector by $\lceil \log_2 T \rceil$ dimensions for
  the second operand, initialized from the direction in
  \texttt{scr5}.

\item \textbf{Adder Chain} (feedforward, Lemma C.2).
  A ripple-carry adder of $\lceil \log_2 T \rceil$
  \texttt{FullAdder} modules computes the new head position
  in \texttt{pos3}.

\item \textbf{Project Down} (linear).
  Drops the extra adder dimensions, returning to width $w$.

\item \textbf{Indicate Visited Position}
  (self-attention + linear, Claim C.4).
  A self-attention layer over $H$ determines whether
  the new position \texttt{pos3} has been visited in any
  previous step.
  The query matches on \texttt{pos3} against the
  \texttt{pos2} of all previous steps.
  A flag is stored in \texttt{scr4}.

\item \textbf{Binary Search}
  ($\lceil \log_2 T \rceil$ self-attention layers, Claim C.5).
  Each \texttt{BinarySearchStep} narrows the search
  for the most recent step that wrote to the current
  position, matching one additional bit of the step number
  per layer.
  The step-number bits of the best match accumulate
  in \texttt{scr3}.

\item \textbf{Get Last Written Symbol}
  (self-attention, Claim C.6).
  Using the step number identified by binary search,
  retrieves the symbol written at that step,
  storing it in \texttt{scr1}.

\item \textbf{Get Initial Symbol}
  (cross-attention, Lemma C.7).
  The single cross-attention layer: the decoder queries
  the encoder $E$ by matching \texttt{pos3} against the
  position indices in $E$, retrieving the original tape
  symbol.
  Result stored in \texttt{scr2}.

\item \textbf{Assemble Symbol}%
  \hspace{0pt plus 3pt}(feedforward, Lemma~C.8).
  Three operations (\texttt{GetV}, \texttt{ArrangeSymbols},
  \texttt{CombineSymbols}) implement a priority:
  (a)~if the position was previously written to, use the
  last written symbol from \texttt{scr1};
  (b)~else if the position is on the original tape, use
  the initial symbol from \texttt{scr2};
  (c)~else use the empty symbol~$E$.
  The result is written to \texttt{sym1} and
  \texttt{pos3} is copied to \texttt{pos2},
  preparing the vector for the next step.
\end{enumerate}

\begin{figure}
\centering
\includegraphics[height=0.55\textheight]{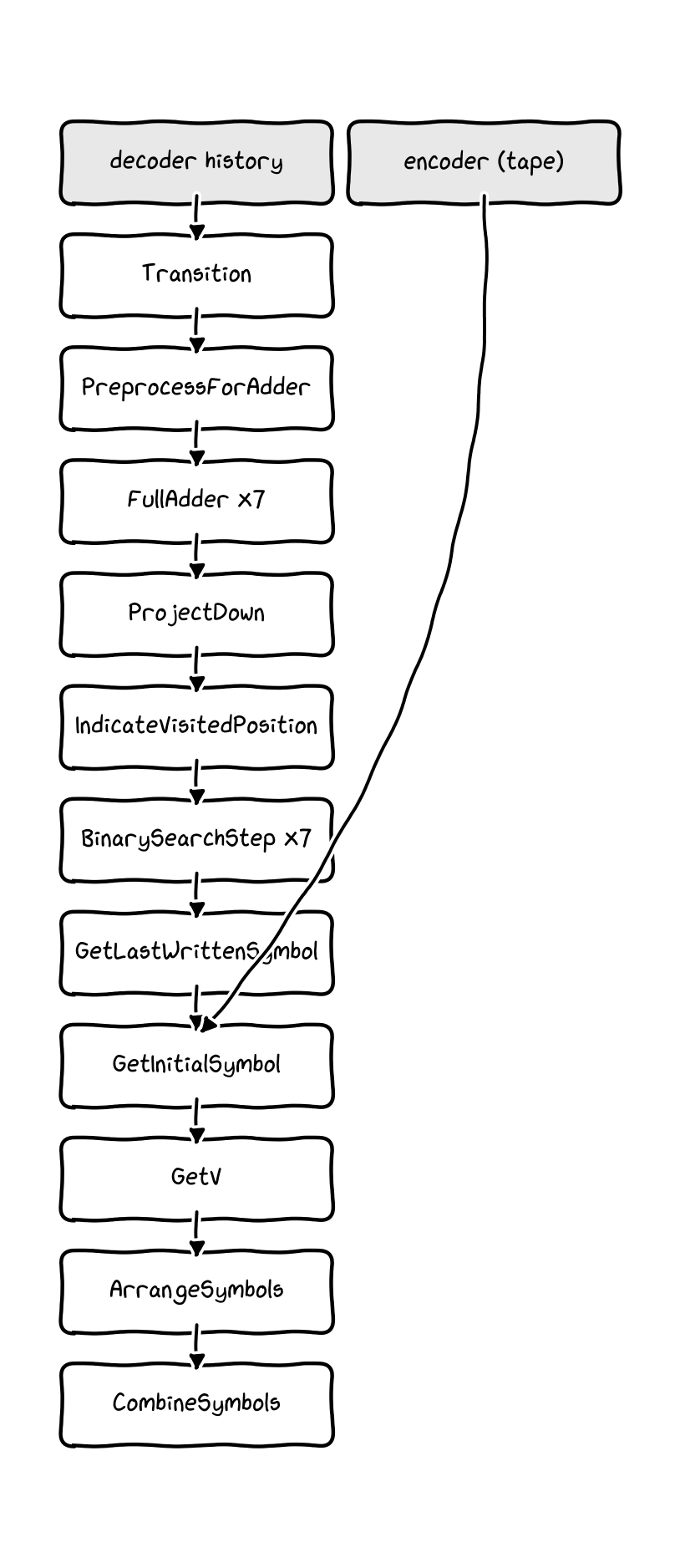}
\caption{Architecture of the WCM21 simulator for one step.
Feedforward (FF) layers implement Boolean circuits;
self-attention layers search the decoder history;
cross-attention reads the original tape from the encoder.}
\label{fig:wcm_pipeline}
\end{figure}

\subsection{Attention Mechanism}

All attention layers use the same functional form.
Given query matrix $Q$, key matrix $K$, and value matrix $V$
(obtained by linear projections with specified weights),
plus a null key $k_0$ and null value $v_0$:
\begin{equation}
\mathrm{Attn}(Q, K, V) = \mathrm{softmax}(\kappa \, QK^\top) V,
\end{equation}
where $\kappa = 9999$ approximates hard attention.
The null key-value pair provides a default output when no
key matches the query.

The query and key projections are designed so that
matching positions produce a high dot product (equal to
the number of matching binary bits), while non-matching
positions score lower.
With the large temperature, softmax concentrates all weight
on the best match, effectively implementing exact lookup.

The network contains three types of attention:
9 self-attention layers (1 for visited-position detection,
$\lceil \log_2 T \rceil$ for binary search, 1 for
symbol retrieval), all attending over the decoder history $H$;
and 1 cross-attention layer, where the decoder queries
the encoder $E$.

\subsection{Complexity}

For a Turing machine with $|Q|$ states, $|\Gamma|$ symbols, and
maximum steps $T$:
the transition layer has $O(|Q|\,|\Gamma|)$ hidden neurons;
the adder chain and binary search each have
$O(\log T)$ layers;
each state vector has width $O(|Q| + |\Gamma| + \log T)$;
and the decoder history grows linearly with $T$.

\section{The SS95 Construction}
\label{sec:ss}

The Siegelmann-Sontag construction~\cite{siegelmann1992,siegelmann1995} takes a
fundamentally different approach: it reformulates the Turing
machine as a \emph{stack machine} and encodes each stack
as a single rational number.

\subsection{Stack Machines}

A $p$-stack machine is equivalent to a Turing machine but
uses $p$ stacks (each supporting push, pop, and peek) instead
of a tape.
For the balanced parentheses problem, two stacks suffice:
stack~0 holds the unprocessed input; stack~1 accumulates
unmatched open parentheses.
The transition function maps
$(\text{state}, \text{top}_0, \text{top}_1)$ to
$(\text{next state}, \text{op}_0, \text{op}_1)$,
where each operation is one of
\texttt{noop}, \texttt{push\,0}, \texttt{push\,1}, or \texttt{pop}.

\subsection{Cantor-Set Encoding}

The key mathematical trick is encoding each stack by a rational number.
For a binary stack (symbols $0$ and $1$) with base $b$ and
parameter $\rho$, the encoding is:
\begin{equation}
\label{eq:cantor}
\mathcal{E}(a_1 a_2 \cdots a_k) = \sum_{i=1}^{k} \frac{b - 1 + 4\rho(a_i - 1)}{b^i}.
\end{equation}
With $b = 4$ and $\rho = 1/2$, symbol $0$ maps to $1/4$
and $1$ maps to $3/4$ of each interval,
corresponding to a Cantor set in $[0, 1]$.
The empty stack maps to $0$.

Note that $a_1$ is the top of the stack and $a_k$, the bottom,
reversed from how stacks are written in Python.

\textbf{push, pop, and peek are linear functions}
of the encoded value.
With $b = 4$ and $\rho = 1/2$,
\begin{align}
\mathrm{push}(v, a) &= \frac{v + 2a + 1}{4}, \\
\mathrm{pop}(v) &= 4v - 2\,\mathrm{top}(v) - 1, \\
\mathrm{top}(v) &= \begin{cases} 0 & \text{if } \sat(2v - 1) = 0, \\ 1 & \text{otherwise.} \end{cases}
\end{align}
Since these are linear or piecewise-linear, a network with
the saturated activation $\sat(x) = \clamp(x,\, 0,\, 1)$
can manipulate stacks using only matrix multiplications
and element-wise clamping.

\begin{figure}[H]
\centering
\includegraphics[width=0.6\textwidth]{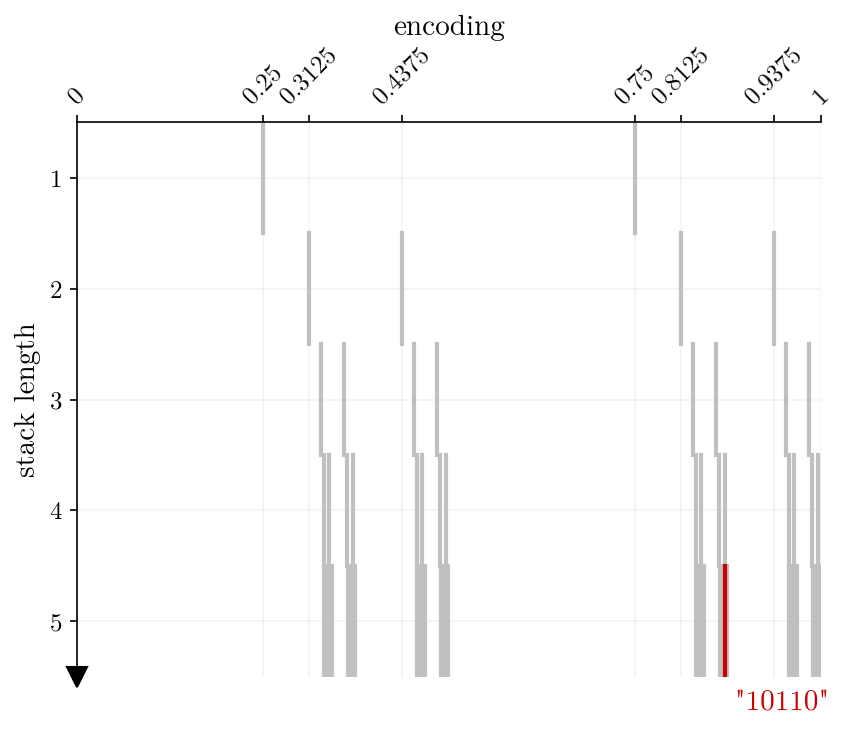}
\caption{Cantor set encodings by stack length.
Each stack symbol selects a subinterval, and the
stack is represented by a single rational number in $[0, 1]$.
Push divides by $b$; pop multiplies by $b$.}
\label{fig:cantor}
\end{figure}

\subsection{Network Architecture}

The network maintains a state vector containing the one-hot
machine state and the Cantor-encoded value of each stack.
At each step, a \emph{configuration detector} identifies
the matching transition rule from the current state and
the top of each stack, then applies the corresponding
stack operations.

The weights connecting the configuration detector to the
output layer are determined by solving a linear system.
Let $d \in \mathbb{R}^{|Q| \cdot 3^p}$ be the detector output
(one entry per state--stack-top combination).
The next-state vector is $\beta\, d$ and the stack-operation
vector is $\gamma\, d$, where the matrices
$\beta$ and $\gamma$ are obtained by solving least squares
against the desired outputs for all possible configurations.

Two versions are implemented:

\paragraph{4-layer version (\texttt{version=4}).}
Each Turing machine step uses 4 network layers in an
explicit pipeline:
layer~F4 reads the configuration (state and stack tops),
F3 detects which transition rule matches,
F2 applies the transition (next state and stack operations),
and F1 reassembles the output vector.
Uses $b = 4$, $\rho = 1/2$, and \texttt{float32} arithmetic.

\paragraph{1-layer version (\texttt{version=1}).}
A single recurrent layer per step,
following the original construction more closely.
Uses $b = 10p^2$, giving $b = 40$ for $p = 2$.
Using a \texttt{float64} stack,
which has 53 bits of mantissa,
each pop amplifies rounding error by a factor of $b$.
After $k$ pops the accumulated error is
$\sim b^k \cdot 2^{-53}$, giving
$\lfloor 53 / \log_2 40 \rfloor = 9$ reliable pops.
This is enough for balanced-parenthesis inputs of length~8
(4~matched pairs).

\section{Implementation}
\label{sec:impl}

The package is implemented in Python using PyTorch~\cite{pytorch}.
The interface requires only a transition function and
terminal states:

\begin{lstlisting}[caption={Simulating with the WCM21 transformer.},label={lst:wcm}]
from turing.wcm.simulator import Description, Simulator

desc = Description(transition_function, terminal_states)
sim = Simulator(desc, T=100)
result = sim.simulate("B()((()(()))())E")
# prints each step; returns "T" (balanced) or "F"
\end{lstlisting}

\begin{lstlisting}[caption={Simulating with the SS95 stack machine.},label={lst:ss}]
from turing.ss.simulator import Description, Simulator

desc = Description(delta_stack, terminal_states)
sim4 = Simulator(desc, version=4)
sim4.simulate("(()())", T=12)

sim1 = Simulator(desc, version=1)
sim1.simulate("(())", T=12)
\end{lstlisting}

Both simulators produce standard \texttt{torch.nn.Module}
objects.
The WCM21 model for the balanced parentheses example
(6 states, 5 symbols, $T = 100$) has embedding width
$w = 59$ and consists of:
a \texttt{Transition} module ($59 \to 89 \to 59$),
a \texttt{PreprocessForAdder} ($59 \to 66$),
7 \texttt{FullAdder} modules (each with 2 \texttt{HalfAdder}
submodules and 2 OR layers),
a \texttt{ProjectDown} ($66 \to 59$),
9 self-attention layers (1 for position detection,
7 for binary search, 1 for symbol retrieval),
1 cross-attention layer (for reading the original tape),
and 3 feedforward layers for symbol assembly.
All feedforward layers use $\relu$; all attention layers
use softmax with $\kappa = 9999$.

Interactive Jupyter notebooks are included:
\texttt{part1} builds the transition layer and binary adder
from scratch;
\texttt{part2} demonstrates the full pipeline with attention.

\section{Discussion}
\label{sec:discussion}

\paragraph{Interpretability and permutation symmetry.}
These networks are interpretable by construction: every weight
has a known function.
This interpretability does not transfer to networks trained
by gradient descent.
A trained network implementing the same function may have
its neurons permuted, rescaled, and entangled.
With $H$ hidden neurons, that's $H!$ permutations, placing
a fundamental limit on post-hoc mechanistic interpretability.

\paragraph{Numerical precision.}
The WCM21 construction operates on binary values in $\{0,1\}$
and is numerically exact in \texttt{float32}.
The SS95 construction is more delicate: the Cantor encoding
stores information in increasingly small fractional digits,
and pop amplifies error by a factor of $b$ per step.
The 4-layer variant ($b = 4$) is stable for typical inputs;
the 1-layer variant ($b = 40$) requires \texttt{float64}
and degrades after approximately 8 pops.
This precision-computability tradeoff is inherent to
Siegelmann-Sontag and illustrates the gap between theoretical
Turing completeness and practical computation.

\paragraph{Future work.}
As standard PyTorch modules, the models can serve as
initializations for gradient-based optimization.
Open questions include:
How flat is the loss landscape around the constructed solution,
and how much weight perturbation can the simulation tolerate?
Does fine-tuning on specific input distributions yield
shortcuts that the general construction misses?

\section{Conclusion}

We have presented a constructive approach to compiling Turing
machines into neural network weights.
Starting from Boolean gates in ReLU networks, we built up
through DNF realization, binary adders, and hard attention
to obtain two Turing machine simulators: a transformer (WCM21)
and a recurrent network (SS95).
The resulting PyTorch package takes a Turing machine
description and produces a model that is correct by
construction, with no training required.
Code is available at
\url{https://github.com/jonrbates/turing}.


\end{document}